\documentclass{edm_article}
\usepackage{graphicx}
\usepackage{enumitem}
\usepackage{url}
\begin{document}

\title{Automated Identification of Logical Errors in Programs: Advancing Scalable Analysis of Student Misconceptions}

% Submissions for EDM are double-blind: please do not include any author names or affiliations in the submission. 
% Anonymous authors:
\numberofauthors{6}
\author{
\alignauthor
Muntasir Hoq\\
       \affaddr{NC State University}\\
       % \affaddr{Raleigh, NC, USA}\\
       \email{mhoq@ncsu.edu}
\alignauthor
Ananya Rao\\
\affaddr{NC State University}\\
       % \affaddr{Raleigh, NC, USA}\\
       \email{arrao3@ncsu.edu}
\alignauthor
Reisha Jaishankar\\
\affaddr{NC State University}\\
       % \affaddr{Raleigh, NC, USA}\\
       \email{rjaisha@ncsu.edu}
\and
\alignauthor
Krish Piryani\\
\affaddr{NC State University}\\
       % \affaddr{Raleigh, NC, USA}\\
       \email{kopiryan@ncsu.edu}
\alignauthor
Nithya Janapati\\
\affaddr{NC State University}\\
       % \affaddr{Raleigh, NC, USA}\\
       \email{nsjanapa@ncsu.edu}
\alignauthor
Jessica Vandenberg\\
\affaddr{NC State University}\\
       % \affaddr{Raleigh, NC, USA}\\
       \email{jvanden2@ncsu.edu}
\and
\alignauthor
Bradford Mott\\
\affaddr{NC State University}\\
       % \affaddr{Raleigh, NC, USA}\\
       \email{bwmott@ncsu.edu}
\alignauthor
Narges Norouzi\\
       \affaddr{UC Berkeley}\\
       % \affaddr{Berkeley, CA, USA}\\
       \email{norouzi@berkeley.edu}
\alignauthor
James Lester\\
\affaddr{NC State University}\\
       % \affaddr{Raleigh, NC, USA}\\
       \email{lester@ncsu.edu}
\and
\alignauthor
Bita Akram\\
       \affaddr{NC State University}\\
       % \affaddr{Raleigh, NC, USA}\\
       \email{bakram@ncsu.edu}
% \and
% \alignauthor
% Anonymous\\
%        \affaddr{Anonymous Institution}\\
%        %\affaddr{City, State, Country}\\
%        %\email{anon@anonymous.edu}
% \alignauthor
% Anonymous\\
%        \affaddr{Anonymous Institution}\\
%        %\affaddr{City, State, Country}\\
%        %\email{anon@anonymous.edu}
% \alignauthor
% Anonymous\\
%        \affaddr{Anonymous Institution}\\
%        %\affaddr{City, State, Country}\\
%        %\email{anon@anonymous.edu}
}

\maketitle

\begin{abstract}
In Computer Science (CS) education, understanding factors contributing to students' programming difficulties is crucial for effective learning support. By identifying specific issues students face, educators can provide targeted assistance to help them overcome obstacles and improve learning outcomes. While identifying sources of struggle, such as misconceptions, in real-time can be challenging in current educational practices, analyzing logical errors in students' code can offer valuable insights. This paper presents a scalable framework for automatically detecting logical errors in students' programming solutions. Our framework is based on an explainable Abstract Syntax Tree (AST) embedding model, the Subtree-based Attention Neural Network (SANN), that identifies the structural components of programs containing logical errors. We conducted a series of experiments to evaluate its effectiveness, and the results suggest that our framework can accurately capture students' logical errors and, more importantly, provide us with deeper insights into their learning processes, offering a valuable tool for enhancing programming education. 

%Logical errors can stem from various issues with student learning and problem-solving processes, and misconceptions have been identified as the most common source of logical errors. We select some of the common underlying sources of logical errors in programming from CS education literature and use expert opinion to categorize logical errors with their potential underlying causes.  We then compare our model's effectiveness across all categories to show its effectiveness in identifying various types of logical errors. 

\end{abstract}

 % Replace with your own 3-5 keywords

\section{Introduction}
To effectively support students in mastering introductory programming, timely identification and resolution of factors contributing to students’ programming challenges, especially difficulties with fundamental concepts, are imperative~\cite{marwan2022adaptive}. However, the increased number of CS majors has created significant challenges in supporting students with limited teaching resources~\cite{hovey2019survey}. As such, instructors' real-time monitoring of students' learning progress is often infeasible~\cite{martin2022intelligent,raigoza2017study}. Addressing this challenge requires an automated approach to efficiently identifying students' programming struggles, enabling personalized feedback delivery through either teachers or automated systems. However, automated identification of students' programming struggles is complex due to the latent nature of learning difficulties, which cannot be directly assessed. Nevertheless, these struggles can manifest as logical errors in student code~\cite{albrecht2020sometimes,ettles2018common}. Computing education research has identified several potential underlying causes for logical errors, including strategic or algorithmic errors, misinterpretation of the problem description, carelessness, and foundational misconceptions~\cite{ettles2018common,qian2017students} with misconceptions noted as the most common source of logical errors~\cite{ettles2018common}.  Thus, analyzing logical error patterns over time can predict students' learning outcomes. More importantly, this analysis can shed light on potential misconceptions held by students, enabling scalable approaches for remediating through adaptive scaffolding~\cite{ettles2018common}. %In this study, we propose a validated approach for the automated identification of logical errors in students' programs and demonstrate their connection with students' learning outcomes. We further discuss well-established adaptive instructional approaches to leverage logical errors to address students' potential misconceptions. 

% Previous research has explored semi-automated approaches for uncovering misconceptions by leveraging association rules, as demonstrated in studies such as \cite{elmadani2012data,guzman2010data}. However, these efforts primarily focused on simplistic association rules, proving insufficient for analyzing more intricate tasks like programming \cite{shi2021toward}. A recent study \cite{shi2021toward} employed deep learning models for automatic misconception identification. Nonetheless, their methodology necessitated manual data labeling using an expert-designed rubric. Additionally, their approach targeted the detection of misconceptions on a per-rubric-item basis within student code, lacking effectiveness in addressing misconceptions that may extend across multiple rubric items or arise in scenarios where rubric items and test cases are absent. However, this suggests that deep learning techniques have the potential to extend beyond mere assessment and contribute to the identification of misconceptions in an automated. This will facilitate timely feedback for any classroom size by aiding teachers \cite{demszky2023can} and students \cite{rivers2013automatic}.

Previous research has explored semi-automated methods for identifying errors and misconceptions using association rules in constrained domains \cite{elmadani2012data,guzman2010data,bian2018nar}; however, these approaches have proven inadequate for complex, general-purpose tasks, such as programming~\cite{shi2021toward}. More recent efforts have applied Deep Learning (DL) models for automatic logical error identification. However, they require manual data labeling via expert-designed rubrics and test cases and focus only on conceptual errors associated with specific rubric items~\cite{shi2021toward}. These approaches may not detect logical errors spanning multiple rubric items or be applicable in scenarios where rubrics or test cases are not readily available. Recent studies~\cite{fwa2024experience,lee2024improving} have explored using Large Language Models (LLMs) to identify logical errors and misconceptions in student programs, but require expert supervision~\cite{jia2024assessing,xu2024hallucination}. Regardless, DL-based approaches show promise in automating logical error detection, providing timely feedback to teachers~\cite{demszky2023can} and students~\cite{rivers2013automatic}, even in large classrooms.
%Our proposed approach leverages state-of-the-art Machine Learning (ML) methodologies to dynamically find traces of potential misconceptions in students' programming and investigate their relationship with students' learning outcomes. The minimal manual labor required for implementing this approach and its ability to dynamically identify unseen logical errors and their potential underlying misconceptions position it as an effective way to scale adaptive instructional support to introductory programming students. 

This paper proposes an automated framework for identifying students' logical errors within their programs using a DL framework. In this work, we utilized a publicly available dataset collected from an introductory programming course at a university in the United States~\cite{edwards2017codeworkout}. Although the programs in this dataset were automatically labeled as correct or incorrect by executing the programs using test cases, we do not utilize these test cases or any rubric to pinpoint logical errors in incorrect student code. Instead, our framework utilizes a modified version of the state-of-the-art explainable DL model, Subtree-based Attention Neural Network (SANN)~\cite{hoq2023sann}, which captures students' programming patterns by embedding program vectors based solely on program correctness information. We modify the attention layer of SANN, which highlights important sub-structures within program snippets, aiding in classifying code as correct or incorrect. Evaluation results indicate that these highlighted patterns closely correlate with logical errors in student code, with our method effectively identifying logical errors in students' incorrect programs. We further engage expert annotators to categorize identified logical errors as one of the two essential categories identified in the literature: \textit{algorithmic error} (strategic) and \textit{fundamental conceptual error} (misconceptions)~\cite{ettles2018common,qian2017students}, and a third category, \textit{compilable syntactic error} that we encountered during our analysis (details in Section~\ref{sec:logic_error_categories}), to better understand the relationship between logical errors and underlying students' struggles. Literature has also referred to \textit{misinterpretation} and \textit{carelessness} as other possible underlying causes for logical errors. However, distinguishing between issues arising from misinterpretation or carelessness rather than conceptual misunderstanding or strategic errors requires internal knowledge about a student's state of mind~\cite{albrecht2020sometimes}. Therefore, we excluded \textit{misinterpretation} and \textit{carelessness} as possible underlying causes for logical errors in this paper.

Misconceptions have been identified as the most common source of logical errors~\cite{ettles2018common}. Simultaneously, research has shown that incorporating misconception information can significantly enhance student modeling results~\cite{liu2016modeling}. Thus, we hypothesize that enriching code vectors with logical error information can shed light on students' learning progress. To verify this hypothesis, we embed the logical error information in program vectors and utilize them in Long Short-Term Memory-based Deep Knowledge Tracing (LSTM-DKT) models to predict students' short-term (next problem attempt results) and long-term (final exam grades) performance. Our results demonstrate improved predictive student modeling and potential for advancing the modeling of student progress. 
% . Incorporating misconception information is an effective way to improve student models \cite{liu2016modeling}.
This framework can also assist in dynamically finding traces of logical errors and potential misconceptions in students' programming, supporting the investigation of their relationship with students' learning outcomes. 

The main contributions of this work are as follows:
\begin{itemize}[topsep=0pt,parsep=0pt]
    \item Developing a novel, explainable deep learning framework that identifies logical errors in student code without relying on external rubrics or test cases.
    \item Validating the framework through expert annotation, categorizing them into syntactic, strategic, and conceptual errors to rigorously assess its effectiveness.
    \item Demonstrating improved predictive student modeling by incorporating logical error information into code embeddings.
\end{itemize}

We further discuss the framework's potential to enhance students' learning by enabling scalable, dynamic, and adaptive scaffolding for students while learning to program. 

%In this study, we propose a validated approach for the automated identification of logical errors in students' programs and demonstrate their connection with students' learning outcomes. We further discuss well-established adaptive instructional approaches to leverage logical errors to address students' potential misconceptions.

% 
%misconceptions~\cite{emerson2020cluster, shi2021more} 

% The experimental results suggest that our proposed framework provides a method for extracting students' logical errors from incorrect programs without expert rubrics or test cases. 

% Additionally, it contributes to learning analytics by offering a data-driven approach to understanding and addressing common errors, ultimately improving educational outcomes in programming courses.

% 
\section{Related Work}
In this section, we discuss related work on logical errors in student code within computer science education, methods for identifying students' logical errors, and predictive student modeling for predicting student learning.

% Knowledge category
\subsection{Logical Errors in CS Education}\label{sec:logic_error_categories}
Various categorizations of programming errors have been proposed in the literature~\cite{ko2003development,albrecht2020sometimes}, with common classifications categorizing errors into syntactic, semantic (logical), and runtime errors~\cite{hristova2003identifying}. Logical errors are prevalent in students' programs and take the most time to identify and resolve due to a lack of direct external feedback, such as ones provided by compilers for syntactic errors~\cite{fitzgerald2008debugging,altadmri201537,alzahrani2021common}. Nevertheless, identifying logical errors can provide invaluable insights into students' struggles with learning to program, including their misconceptions about programming concepts~\cite{ettles2018common}.

Prior research in CS education has identified various sources for logical errors, including \textit{conceptual}, \textit{strategic}, \textit{misinterpretation}, and \textit{carelessness}~\cite{albrecht2020sometimes,ettles2018common}.
\begin{itemize}[topsep=0pt]%,parsep=0pt]
    % \item Syntactic Knowledge: Referring to the syntax or rules of a programming language, syntactic knowledge encompasses errors such as using commas instead of semicolons in a for-loop header or utilizing undeclared variables.
    \item \textit{\textbf{Conceptual errors}}: A conceptual error is a logical error involving a misunderstanding of how a programming construct works, reflecting a fundamental flaw in programming knowledge. An example of a conceptual error is when a student assumes that a variable retains its previous value after being reassigned.
    \item \textit{\textbf{Strategic errors}}: A strategic error occurs when a student's underlying algorithm or problem-solving approach is flawed, even if their understanding of programming constructs is sound. For example, failing to correctly handle the base case in recursion.% Strategic errors represent higher-level mistakes in planning or problem-solving and do not necessarily indicate a lack of programming knowledge. Instead, they reflect a misstep in the logical design of the solution before the actual programming task begins.
    % \item Sloppiness: Referring to unintentional errors such as forgetting a semicolon or typos.
    \item \textit{\textbf{Misinterpretation errors}}: A misinterpretation error occurs when a student interprets a programming task or question differently than intended, leading to discrepancies between the expected and actual program output.
    \item \textit{\textbf{Carelessness errors}}: A carelessness error is an unintentional mistake resulting from oversight or inattentiveness rather than a deeper misunderstanding. Common examples include missing punctuation (e.g., semicolons) or typographical errors (e.g., using $=$ instead of $==$) that do not stem from conceptual issues.
\end{itemize}

In this paper, we focus on two categories of logical errors from the literature: \textit{conceptual} and \textit{strategic} errors, since the other categories (\textit{misinterpretation} and \textit{carelessness}) cannot be readily inferred from student code due to their inherently latent nature and do not necessarily indicate a lack of programming knowledge~\cite{albrecht2020sometimes}. Although we did not include identifying programming language-specific syntactic errors from uncompilable student submissions in this work that can be easily detected at compile time~\cite{ettles2018common}, we encountered a third category identified in a small percentage of student programs during our analysis, which we classify as \textit{compilable syntactic} errors. For example, using \texttt{\&} instead of \texttt{\&\&} in a conditional statement, which some compilers (e.g., Java) take as a bitwise operation and compile the code. This type of error can be language-specific and even stem from either \textit{carelessness} or \textit{conceptual} errors.

% Previous misconception works
\subsection{Logical Error Detection}
Misconceptions in CS education refer to incorrect interpretations of particular concepts, leading to logical errors in programming~\cite{swidan2018programming,ettles2018common}. Despite the prevalence of logical errors among novice learners, these misunderstandings often evade detection by experts and instructors~\cite{sorva2013notional}, a phenomenon known as the ``expert blind spot''~\cite{nathan2003expert}. Identifying and comprehending these logical errors increases instructors' awareness and provides essential material for modeling students' learning over time to improve learning outcomes~\cite{liu2016blending,liang2022help}. Although previous research has explored automated methods for detecting student logical errors~\cite{guzman2010data,ettles2018common,shi2021toward}, many of these approaches require expert labeling of training data, presupposing familiarity with the specific logical errors being targeted.

In CS education, identifying logical errors and misconceptions traditionally relies on labor-intensive manual analysis of students' programs~\cite{paul2013hunting} or interviews~\cite{kaczmarczyk2010identifying}. 
%Many studies on programming errors have focused on analyzing compiler messages due to their ease of automatic evaluation~\cite{brown2014blackbox,denny2012all,jackson2005identifying,jadud2005first}. They only dealt with syntactical and compile-time errors~\cite{albrecht2020sometimes}. 
The manual construction of ``bug libraries'' can also pose challenges for experts due to its time-consuming nature~\cite{baffes1994learning,davies2015using,ardimento2020software}. Nonetheless, such manual analyses, particularly those conducted on a series of program snapshots, have been shown to offer deeper insights into students' misconceptions. For instance, Albrecht and Grabowski~\cite{albrecht2020sometimes} manually inspected students' programming data and classified errors into six categories: \textit{syntactic}, \textit{strategic}, \textit{conceptual}, \textit{carelessness}, \textit{misinterpretation}, and \textit{domain} knowledge.

Other research has explored data-driven approaches to uncovering logical errors. For example, Guzman et al.~\cite{guzman2010data} used association rule mining to identify misconceptions from closed-ended multiple-choice questions. However, this method relies on experts to label rules and does not target open-ended programming problems. Ettles et al.~\cite{ettles2018common} grouped common logical errors from students' incorrect programs using expert-designed test cases. A recent study~\cite{shi2021toward} employed DL models for automatic misconception identification. Nonetheless, their methodology necessitated manual data labeling using an expert-designed rubric. Additionally, their approach targeted the detection of conceptual logic errors on a per-rubric-item basis within student code, lacking effectiveness in addressing logical errors that may extend across multiple rubric items or arise in scenarios where rubric items and test cases are absent. In some recent studies, LLMs were used to identify logical errors and misconceptions in students' programs~\cite{fwa2024experience,lee2024improving}. However, using LLMs without expert supervision and explainability might suffer from hallucination, precision, and stakeholder safety~\cite{jia2024assessing,xu2024hallucination}.

\subsection{Student Modeling}
Student modeling is crucial in understanding and tracking students’ mastery of skills for specific tasks within the learning paradigm~\cite{karlovvcec2012knowledge}. Additionally, predicting students’ performance and success is pivotal in adaptive learning environments designed to provide adaptive scaffolding~\cite{hoq2023analysis}. Prior studies have focused on various objectives of performance prediction: short-term performance, such as next problem success prediction by knowledge tracing~\cite{yang2022code}, and long-term performance, such as final exam grade prediction from students’ programming submissions~\cite{yoder2022exploring}. Research indicates that incorporating misconceptions can enhance student models \cite{liu2016modeling}, particularly given that misconceptions are a common source of logical errors \cite{ettles2018common}. Thus, code vectors enriched with such logical error information may offer deeper insights into students’ learning trajectories and outcomes.

% knowledge tracing
Knowledge tracing (KT) models student knowledge dynamically as students engage with problems, enabling predictions about future performance in students’ problem-solving attempts. In KT, problems are typically associated with required skills, and the relationship between problems and skills can either be learned from data or derived directly from problem IDs~\cite{yang2022code}. KT models can predict students’ success in future attempts by keeping track of the mastery of the required skills. Bayesian Knowledge Tracing (BKT)~\cite{corbett1994knowledge} is one of the most widely used KT models, and it has undergone several enhancements over the years~\cite{baker2008more,yudelson2013individualized}. A significant advance was introduced by Piech et al. \cite{piech2015deep}, who proposed Deep Knowledge Tracing (DKT). This model utilizes recurrent neural networks (RNNs) to predict a student’s knowledge of each skill after each problem attempt. 
%while automatically learning the interrelationships among skills.
Despite these advancements, existing KT models have limitations when applied to programming data, particularly when incorporating code-related information. Recent iterations of DKT have attempted to address this gap by integrating code information~\cite{yang2022code,liang2022help,jia2023attentive,shi2024evaluating,wang2024deep}. However, these approaches often require experts to manually label errors/concepts, which can be resource-intensive. Furthermore, they often lack the interpretability to explain the failure prediction by mapping it back to the logical errors of students.

%final exam performance prediction
To track students’ long-term learning progression and learning outcomes, researchers have predicted a range of outcomes, such as predicting student performance in final exams~\cite{hoq2023analysis}, detecting failing students~\cite{1_jamjoom2021early}, and identifying early dropouts~\cite{glandorf2024temporal}. A comprehensive review conducted by Shahiri et al.~\cite{3_shahiri2015review} demonstrated the reliance of many studies on grading-based features such as cumulative grade point averages (CGPA) and intermediate assessment scores (e.g., quizzes, midterms) for predicting course performance. In another study by Jamjoom et al., ~\cite{1_jamjoom2021early}, decision tree and SVM were utilized to categorize enrolled students into passing and failing categories based on features such as quiz and midterm exam scores, with the aim of early intervention for at-risk students. A recent study~\cite{hoq2024explaining} utilized student programming submission data to predict students' final exam grades in an explainable manner, aiming to explain the relationship between students' submission data and final exam grades. However, none of these studies incorporated direct programming code information into the prediction task to integrate students' programming knowledge information for predicting students' performance. In recent work, DL frameworks have become increasingly prominent. In some instances, AST-based and control flow graph-based embedding models were employed to predict students' final exam grades from programming assignment data~\cite{yoder2022exploring,anon_3}.

Despite various efforts to detect logical errors in student code, most prior approaches rely heavily on expert-generated test cases and rubrics or lack explainability and face reliability concerns. In contrast, our work introduces a framework for automated, explainable, and scalable identification of logical errors from students' programs without the need for external rubrics or test cases. We further show that these program snippets embedded in code vectors, many of which represent misconceptions in students' programming knowledge, can be used in different predictive modeling tasks. We propose utilizing these program snippets directly to inform the provision of adaptive scaffolding based on vector similarity measurements as a viable solution to address potential misconceptions in students' programming knowledge and skills.

\section{Dataset}
We use a publicly available dataset\footnote{https://pslcdatashop.web.cmu.edu/Files?datasetId=3458} that was sourced from the CodeWorkout platform\footnote{https://codeworkout.cs.vt.edu/}~\cite{edwards2017codeworkout} used in an introductory programming course at a public university in the United States. This dataset consists of anonymized student programming solutions in Java for 50 programming problems from five assignments used in a CS1 course in the Spring 2019 semester, without any demographic information. The submissions were evaluated on a $[0, 1]$ scale based on the number of passed test cases. We group the data into binary classes of correct and incorrect (if any of the test cases fails) solutions for model training. The dataset also consists of students' final exam grades normalized on a 0 to 1 scale. Throughout the semester, $57,670$ student code submissions were received from $368$ students for $50$ programming problems, comprising $18,787$ correct submissions and $38,883$ incorrect submissions. In this study, we leverage the Abstract Syntax Trees (ASTs) of programs to capture programming patterns. Consequently, consistent with prior research~\cite{shi2021more,shi2021toward}, $9,906$ incorrect uncompilable solutions that could not be parsed into ASTs using our Java parser (javalang\footnote{https://github.com/c2nes/javalang}) were excluded from the dataset.
%$18,873$

\section{Methodology}
 This section describes our adaptations to SANN for effectively identifying students' logical errors and the evaluation process we utilized. We then describe our processes for predicting students' short-term (next problem attempt) and long-term (final exam grade) success.
%  %leverage SANN~\cite{hoq2023sann}, a state-of-the-art explainable code representational DL framework. We 
% modify the SANN's original architecture to effectively identify students' programming errors. % We use the model's attention mechanism to identify logical errors from students' incorrect code submissions. We then used expert-identified logical errors to evaluate the effectiveness of our approach for incorrect student submissions. % in one problem. We extended the evaluation to subsets of incorrect submissions from four different problems. 
% %We further use expert opinion to categorize logical errors into two important categories of logical errors from the literature: conceptual, and strategic ~\cite{bayman1988using}. We then evaluate SANN's ability on picking the logical errors in each of these categories.
% To demonstrate the connection between these explainable logical-error-information-enriched code vectors and students' modeling, we capture students' learning progression throughout the semester by predicting the next-attempt success in solving problems and by their final exam grades.

\subsection{Extracting Logical Errors with SANN}

% \begin{figure*}[!ht]
%   \centering
%   % First subfigure: 70% width
%   \begin{subfigure}{0.64\textwidth}
%     \centering
%     \includegraphics[width=\textwidth]{Figures/sann.jpg}
%     \caption{Model architecture}
%     \label{fig:sann}
%   \end{subfigure}
%   \hfill
%   % Second subfigure: 30% width
%   \begin{subfigure}{0.34\textwidth}
%     \centering
%     \includegraphics[width=\textwidth]{Figures/subtree_split.png}
%     \caption{Subtree extraction from Java code AST}
%     \label{fig:subtree_split}
%   \end{subfigure}
  
%   \caption{Modified SANN architecture}
%   \label{fig:combined}
% \end{figure*}

SANN~\cite{hoq2023sann} is known for its explainability and ability to encode programs into condensed vectors by extracting substructures from ASTs. These vectors are versatile and have been utilized in different prediction tasks in the literature (i.e., program correctness prediction, algorithm detection, and LLM-generated code detection)~\cite{hoq2023sann,hoq2024detecting}. To identify logical errors in students' programs, we utilize SANN's attention mechanism to identify the subtrees extracted from program ASTs that are most influential in the prediction. 
%SANN generates vector representations of source code by encoding subtrees extracted from the AST. 
Subtrees are embedded using a two-way embedding approach in SANN, where each subtree and its nodes are separately embedded. SANN merges the embeddings from the two-way embedding approach into a single embedded vector. Subsequently, the embedded vectors from both approaches are concatenated and passed through a time-distributed, fully connected layer
to generate subtree vectors. % The time-distributed fully connected layer generates the subtree vectors. % $sv_i$ from the embedded vector $e_i$ and applies $tanh$ activation function element-wise. % on the multiplication of $e_i$ with the weight matrix $W$. This combines the node-level and subtree-level information.
%\[sv_i = tanh(W\cdot e_i)\]
%For convenience, the height of W in SANN is kept the same as that of $e_i$, which is determined by the subtree vector length. 
Following this, an Attention Neural Network condenses all subtree vectors into a single source code vector. The attention mechanism assigns scalar weights to each subtree vector, supporting the aggregation of all subtree vectors into a weighted average. These weights are determined through a normalized inner product between each subtree vector and a global attention vector, followed by a softmax function. %This weighted average of subtree vectors, determined by the attention mechanism, represents the entire source code snippet. Overall, the SANN model leverages attention weights to prioritize the most significant subtrees when generating the source code vector.

%Using the attention network, we can identify the most important subtrees of a program according to a prediction task. 

\begin{figure*}[h]
  \centering
  \includegraphics[width=0.9\textwidth]{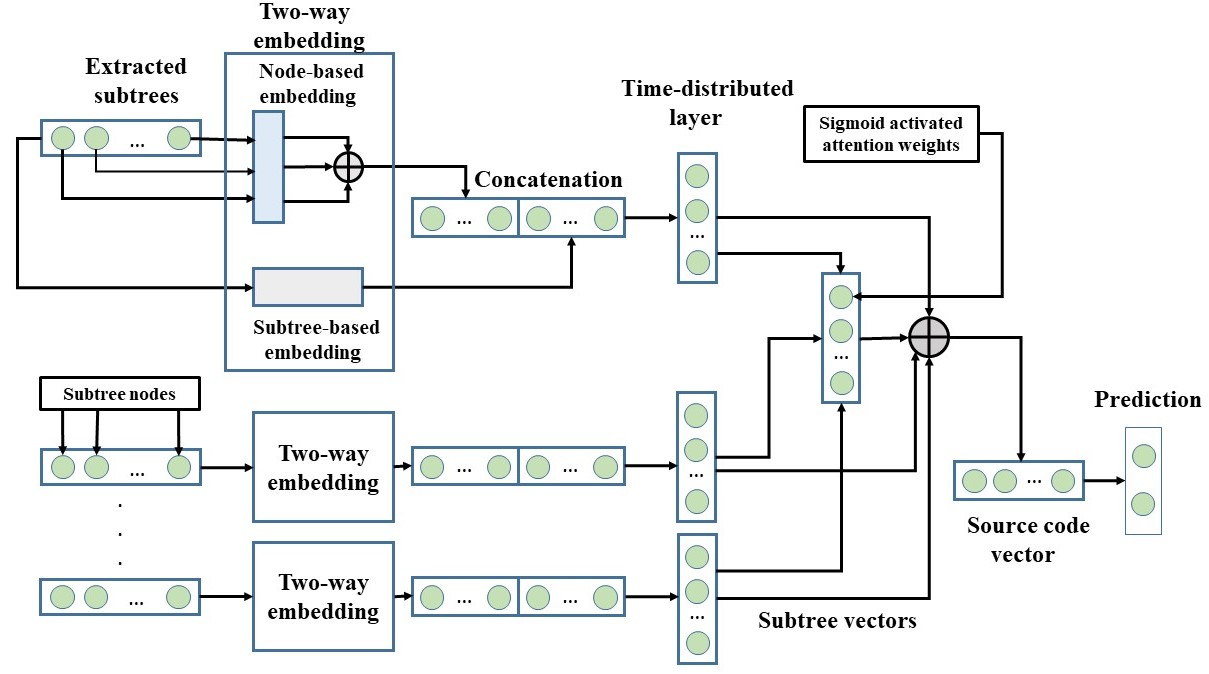}
  \caption{Model architecture of SANN.}
  \label{fig:sann}
  \Description{Architecture of the SANN model with the two-way embedding approach and the attention neural network.}
\end{figure*}

\begin{figure}[h]
  \centering
  \includegraphics[width=\columnwidth]{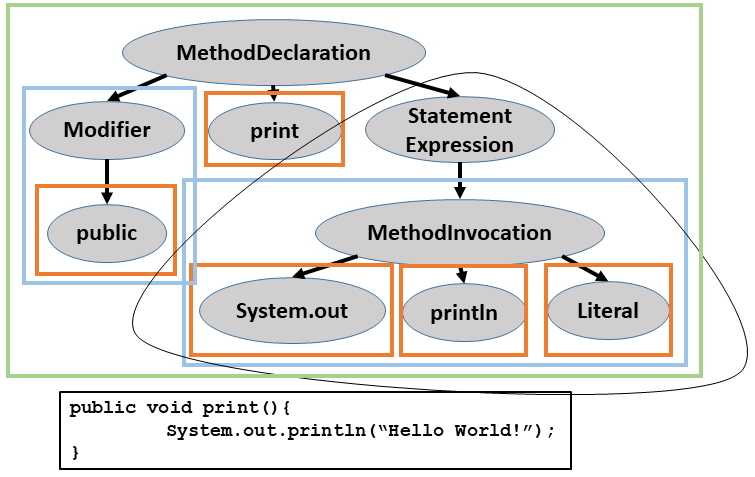}
  \caption{Subtree extraction from an AST.}
  \label{fig:subtree_split}
  \Description{Subtree extraction process of the SANN model where all possible subtrees are extracted from an AST.}
\end{figure}

The original SANN model focuses primarily on only one subtree with the most attention. Typically, this subtree includes an incorrectly implemented part (an error) of the program while predicting it as incorrect. However, we want to identify all logical errors present rather than only one error per incorrect student submission. %In this study, we employ a modified version of SANN to predict the correctness of the student code (0: incorrect, 1: correct) and identify important substructures (subtrees) responsible for the prediction, often representing logical errors in the incorrect code~\cite{hoq2024towards}. 
Figure \ref{fig:sann} illustrates the SANN model architecture. The original version of SANN uses an optimization approach to identify the best strategy for chunking an AST into its substructures for a prediction task. We use a modified version of SANN, where all possible substructures are included in the embedding and training process~\cite{hoq2024towards}. To do this, we extract all subtrees recursively from an AST, as illustrated in Figure \ref{fig:subtree_split}. This enables us to capture logical errors of different granularities. 
%because analyzing code in fixed-sized subtrees might not be sufficient to catch all logical errors. 
Logical errors can vary in size, and a single error might involve interconnected parts of code working together. Some errors might even form a hierarchical structure, with a larger error encompassing a smaller one. 

In addition to modifying the SANN subtree extraction process, the attention mechanism has been enhanced to improve its ability to focus on relevant subtrees. In the original version, attention weights were calculated using a softmax function to understand the importance of different subtrees within the code. However, this approach could inadvertently assign disproportionate attention to a single logical error, thereby increasing the risk of failing to identify multiple logical errors that may simultaneously contribute to the incorrectness of the code. In the modified version, we replaced the softmax function with a sigmoid activation function to compute the attention weights. Unlike softmax, sigmoid activation allows each attention weight to be computed independently without normalizing across all subtrees. This enables the model to assign high attention weights to multiple relevant subtrees and mitigates the risk of assigning excessive attention to a single logical error when multiple logical errors exist in incorrect code. However, based on our initial experimental results, this modification can result in higher false positive logical error predictions. To further refine the focus of the attention mechanism, we introduced an entropy regularization term to the model's loss function to selectively assign attention to fewer subtrees, reducing false positive predictions~\cite{pereyra2017regularizing}.

For each subtree vector $sv_i$, the attention weight $a_i$ is computed as:
\vspace{0.2cm}
\[
a_i = \sigma(sv_i^T \cdot av)
\]
where $\sigma$ is the sigmoid activation function, and $av$ is a global attention vector learned by the model. The entropy regularization term is computed as:
\vspace{0.2cm}
\[
H(a) = -\lambda \sum_{i=1}^{n} a_i \log(a_i + \epsilon)
\]
This entropy regularization loss is added to the overall loss function, along with the prediction loss. Here, $\lambda$ is a regularization weight, $\epsilon$ is a small constant added for numerical stability, and the negative sign ensures entropy is a positive value. By penalizing high entropy, this term encourages the model to avoid distributing attention uniformly across all subtrees, effectively guiding it to focus selectively on a smaller subset of subtrees, thereby improving its ability to identify the logical errors in incorrect code predictions.

This combined approach allows our modified attention mechanism to balance the focus on multiple logical errors, if present, while avoiding assigning unnecessary attention to subtrees that do not contribute meaningfully to the prediction of an incorrect code. We hypothesize that the most important subtrees in an incorrect student code will likely encapsulate the logical errors therein. % By extracting these influential subtrees, we aim to gain insight into students' logical errors and misconceptions. 
Leveraging our modified attention mechanism of the SANN model and extracting these important subtrees, we attempt to identify logical errors within an incorrect submission solely based on the program's correctness without requiring additional information, such as rubrics or test cases utilized in prior studies.

\subsection{Evaluation of Logical Error Identification}

\begin{figure}[h]
  \centering
  \includegraphics[width=\columnwidth]{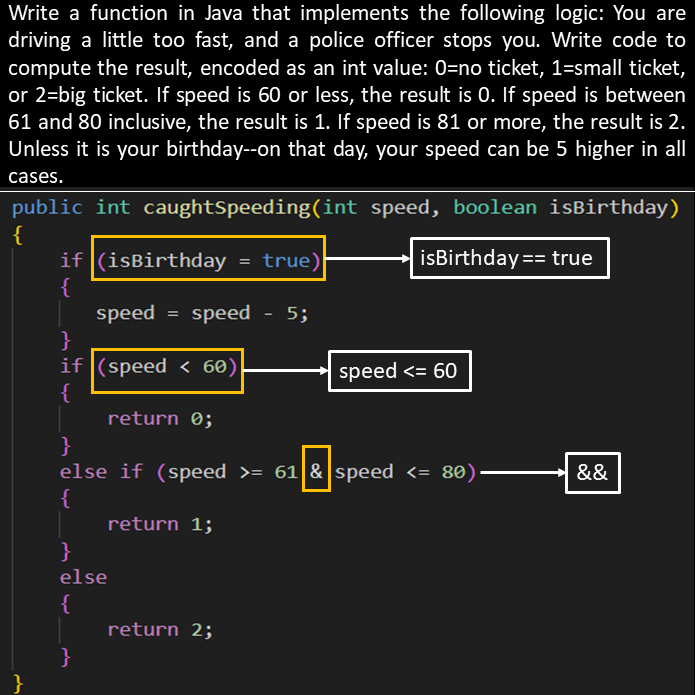}
  \caption{Example incorrect solution for \texttt{caughtSpeeding}.}
  \label{fig:incorrect_code}
  \Description{An incorrect student solution with three logical errors.}
%\vspace{-2mm} % Adjust the value as needed
\end{figure}

To evaluate the effectiveness of our framework in detecting logical errors, two experts labeled all the incorrect submissions from one problem called \texttt{caughtSpeeding} ($1,574$ submissions: $617$ correct, $957$ incorrect) with their logical errors. Figure \ref{fig:incorrect_code} depicts an example of an incorrect student solution to the \texttt{caughtSpeeding} problem with three errors highlighted in orange boxes and their corresponding corrected versions in white boxes. Later, we extended the evaluation to 4 other problems: \texttt{redTicket}, \texttt{countCode}, \texttt{sum13}, and \texttt{canBalance} and selected a subset of incorrect submissions (200 incorrect solutions each) to demonstrate the framework's robustness. 

\begin{figure}
  \centering
  \includegraphics[width=\columnwidth]{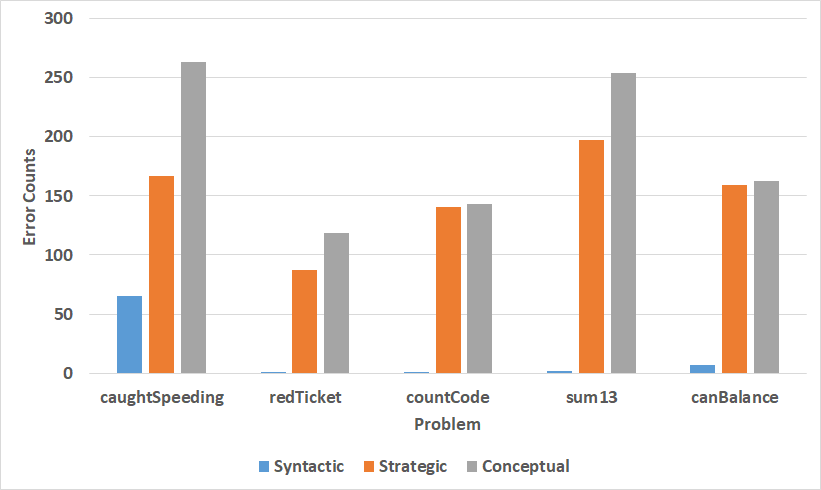}
  \caption{Distribution of logical error types across problems.}
  \label{fig:error_dist}
  \Description{Distribution of different logical errors across problems.}
% \vspace{-2mm} % Adjust the value as needed
\end{figure}

For the first problem, each expert independently labeled $10$\% of the dataset, identifying all existing logical errors in incorrect student submissions. Notably, if the student code contained multiple logical errors and one expert labeled it with fewer errors, it was deemed an incorrect labeling. The labeling process was iterative, with experts calculating Cohen's Kappa scores ($\kappa$) after each round. When the agreement was below $0.8$, they resolved disagreements and continued labeling new data~\cite{landis1977measurement}. The experts achieved $\kappa = 0.78$ in the first round and $\kappa = 0.94$ in the second round. After reaching sufficient agreement, they split the remaining $80$\% of the data, each expert labeling $40$\%. They also labeled each logical error with one of the three potential sources mentioned in Section~\ref{sec:logic_error_categories}. This includes compilable syntactic, strategic, and conceptual errors~\cite{bayman1988using,albrecht2020sometimes,ettles2018common}. The kappa scores of the two rounds for the three categories were Syntactic: $0.84$ and $0.96$, Strategic: $0.72$ and $0.86$, and Conceptual: $0.78$ and $0.92$. The same labeling procedure was followed for the other problems. Figure~\ref{fig:error_dist} shows the distribution of different logical error categories from each problem. Subsequently, the experts then evaluated the effectiveness of the proposed framework in identifying logical errors by comparing them to their own identified logical errors. 

Unlike prior work~\cite{lee2024improving} that relied on accuracy for evaluation, we found it insufficient for measuring the model’s ability to detect all relevant errors while avoiding false positives where the model incorrectly assigns higher attention to a subtree as a logical error. Thus, to evaluate the performance of our framework, we employed both recall and precision. Recall measures the proportion of errors detected by the model compared to the total number of errors identified by expert evaluators. This metric highlights the model's accuracy in capturing the errors that experts considered relevant. On the other hand, precision reflects the proportion of important subtrees identified by the model that truly contain errors, preventing an excess of false positives that could be misleading. Focusing on these metrics provides a more nuanced understanding of the model’s ability to detect meaningful patterns while ensuring that the substructures flagged by the model are genuinely associated with logical errors. 

\subsection{Deep Knowledge Tracing}

Deep knowledge tracing (DKT)
predicts students' short-term performance (next attempt's success), capturing their mastery of skills~\cite{piech2015deep}. It represents a sequence of student attempts, \( S = \{x_1, x_2, \dots, x_T\} \), where each attempt \( x_t \) corresponds to a student's interaction at time \( t \) within a sequence of \( T \) total attempts. Each attempt involves a problem-correctness pair \( \{q_t, a_t\} \), where \( q_t \) denotes the problem identifier and \( a_t \) indicates whether the attempt was correct (1) or incorrect (0). Given a total of \( M \) problems, each attempt \( x_t \) is one-hot encoded into a binary vector of size \( 2M \). The element corresponding to \( x_{q_t + M(1-a_t)} \) is set to~1, while all other entries remain 0. For example, in the case of \( M = 2 \) problems, if a student correctly solves problem~1 (i.e., \( q_t = 1 \) and \( a_t = 1 \)), the corresponding vector becomes \( x = \{1, 0, 0, 0\} \), where \( x_{1 + 2(1-1)} = 1 \). Conversely, if the student fails on problem 1 (i.e., \( q_t = 1 \) and \( a_t = 0 \)), the vector is represented as \( x = \{0, 0, 1, 0\} \), where \( x_{1 + 2(1-0)} = 1 \).

\begin{figure}  % 'r' for right, and width 35% of text width
  \centering
  \includegraphics[width=0.85\columnwidth]{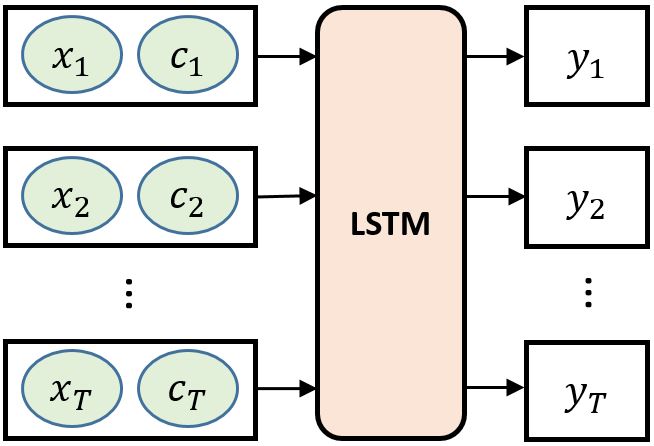}
  \caption{DKT with code vectors.}
  \label{fig:dkt}
  \Description{Architecture of the improved DKT model with code vector integration.}
\end{figure}

We extract code vectors from the modified SANN model, which implicitly encodes logical error information. The explicit logical errors can be retrieved when necessary. In this study, we concatenate code vectors \((c_1, c_2, \dots, c_T)\) with the original DKT~\cite{piech2015deep} input structure to feed into an LSTM model, as shown in Figure \ref{fig:dkt}. The output \( y_t \) is a vector of length equal to the total number of problems, where each entry represents the predicted probability that the student will correctly answer a specific problem. The model's prediction for the correctness of the next attempt, \( a_{t+1} \), can be inferred from the entry in \( y_t \) corresponding to the next problem \( q_{t+1} \).

\subsection{Student Performance Prediction}

\begin{figure}[h]  % 'r' for right, and width 35% of text width
  \centering
  \includegraphics[width=\columnwidth]{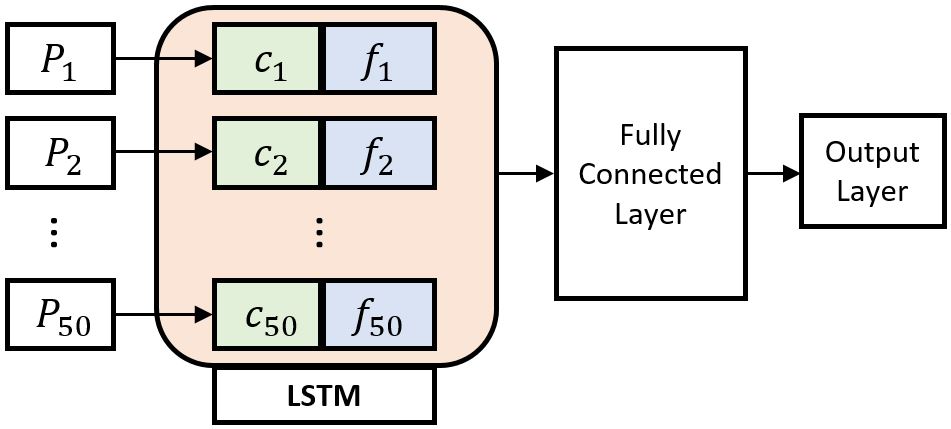}
  \caption{Student performance prediction architecture.}
  \label{fig:lstm}
  \Description{Architecture of the final exam grade prediction model with code vector sequences embedded in an LSTM model to predict the final exam grade.}
\end{figure}

We employ the vector representations derived from students' programming submissions to all 50 problems $(P_1, P_2, \dots, P_{50})$ to forecast their final exam grades in the course. This predictive task offers insights into the embedding capability of our approach, enabling the capture of student programming learning information for student modeling purposes. To conduct this analysis, we adopt a methodology similar to a previous study~\cite{yoder2022exploring}, which utilized code embeddings from the \texttt{code2vec} model to train an LSTM that predicts student performance based on their programming submission sequence. Figure \ref{fig:lstm} shows the architecture of the prediction process.

% \begin{figure}[h]
%   \centering
%   \includegraphics[width=0.35\textwidth]{Figures/lstm.png}
%   \caption{Student performance prediction}
%   \label{fig:lstm}
%   %\Description{Grade distribution of the final exam grades of students with the mean grade.}
% \end{figure}

In our dataset, a student can have multiple submissions for each problem. We use all the code vectors for these submissions to predict students' final exam grades and encode their learning progression. We concatenate all student submissions for a single problem into a vector ($c_i$). Since there was a variable number of submissions for each student-problem pair, we truncated or padded all of the vectors for each submission to create a sequence of length $30$. This length was chosen because, for the training dataset, $99\%$ of student-assignment pairs had $30$ or fewer submissions. The code vectors were pre-padded with vectors of zeros for problem-student pairs where students made fewer than $30$ submissions. For problem-student pairs where students exceeded $30$ submissions, the last $30$ were kept, assuming the latest submissions were more useful.

We also incorporate additional features ($f_i$) for each problem, such as the number of correct and incorrect submissions, the total number of submissions, and the problem ID for each student. These features are obtained by aggregating a student's attempts over all 50 assignments. We then augment students' programs with information related to their problem-solving behavior to predict their final exam grades. We use an LSTM network to predict the final exam grades based on the time series feature set containing student submissions for the $50$ problems and their problem-solving behavior. After the LSTM layer, we have a fully connected layer with \textit{tanh} activation function to incorporate the information from the LSTM layer and improve the model's expressiveness. Finally, we have a dense layer to predict the final exam grade of a student based on the programming submissions.

\section{Experiments and Results}
In this section, we present the experts' evaluation of our framework to assess its effectiveness for identifying logical errors. We further demonstrate our framework's relative effectiveness in picking the logical errors from different categories (i.e., compilable syntactic, strategic, and conceptual). Lastly, we highlight the utility of the code vectors enriched with logical error information in capturing students' learning progressions by presenting the experimental results of student models that predict students' short-term and long-term successes.

\subsection{Logical Error Identification Effectiveness}

Before extracting logical errors from students' incorrect programs, we show the effectiveness of our modified SANN model in predicting program correctness compared to baseline models. We trained the modified SANN model based on the correctness of the student submissions for the $50$ problems (correct: $18,787$, incorrect: $28,977$), framing it as a binary classification task. The dataset was split 80:10:10 for training, validation, and testing in a stratified way due to the imbalanced nature of the dataset. Key training parameters included setting the embedding size of the code vectors to $128$, training over $100$ epochs with an early stopping of $20$ to avoid overfitting, and utilizing the \textit{Adamax} optimizer (learning rate = $0.001$) based on the validation data. The entropy regularization weight was carefully tuned to a minimal value ($3.5 \times 10^{-5}$), ensuring that it did not dominate the primary prediction loss and consequently degrade overall predictive performance. To ensure our modified SANN model can successfully analyze students' code, we conducted comparative evaluations against several existing models from the literature. These models include code2vec~\cite{alon2019code2vec} and ASTNN~\cite{zhang2019novel}, both AST-based code representational models and a few traditional ML models. Table \ref{tab:prediction} shows that our modified SANN model outperformed the other models with the highest accuracy, precision, recall, and F1-score with values of 0.87, 0.86, 0.88, and 0.87, respectively. This demonstrates that our modified SANN model can effectively analyze students' programs.

\begin{table}
\centering
\caption{Performance comparison for program correctness prediction.}
%\vspace{2mm}
\begin{tabular}{|c|c|c|c|c|} \hline
\textbf{Model} & \textbf{Accuracy} & \textbf{Precision} & \textbf{Recall} & \textbf{F1-score}\\ \hline
SVM & 0.74 & 0.75 & 0.63 & 0.63\\ \hline
KNN & 0.75 & 0.75 & 0.66 & 0.70\\ \hline
XGBoost & 0.78 & 0.76 & 0.77 & 0.76 \\ \hline
code2Vec & 0.81 & 0.84 & 0.83 & 0.83\\
\hline
ASTNN & 0.83 & 0.82 & 0.78 & 0.80\\
\hline
SANN & \textbf{0.87} & \textbf{0.86} & \textbf{0.88} & \textbf{0.87}\\ \hline
\end{tabular}
\label{tab:prediction}
\end{table}

To evaluate the effectiveness of our model on logical error identification, we calculated recall for the \texttt{caughtSpeeding} problem through expert evaluation. Our framework identified $96\%$ of the logical errors from all the code (with an attention threshold of $20\%$). Next, we calculated the ability of our framework to identify multiple errors from an incorrect submission. In this evaluation, we marked the identification of a program's logical errors as incorrect if our framework missed at least one of the logical errors present in the code. This means that if there were multiple errors, for example, three, and our framework identified only two of them, we marked it as an incorrect identification. From the expert labels, the maximum number of logical errors in an incorrect program was 10, and the minimum was 1. The average number of logical errors was 2.6. The recall of identifying all logical errors from incorrect solutions using this process was $86\%$. We also examined the logical errors from different categories labeled by the experts. Our model achieved recalls of $100\%$ in identifying compilable syntactic errors, $95\%$ in identifying strategic errors, and $94\%$ in identifying conceptual errors from all the incorrect submissions. Using expert evaluation, we also calculated the precision of the model, which indicates the proportion of important subtrees identified by the model that contain errors. For the \texttt{caughtSpeeding} problem, the precision of the model was calculated to be $90\%$, highlighting its ability to pinpoint the erroneous substructures within the code effectively. This precision metric further suggests the model’s effectiveness in identifying relevant code segments contributing to logical errors.

\begin{table*}
\centering
\caption{Recall (\%) and precision (\%) of the performance on different problems.}
%\vspace{2mm}
\begin{tabular}{|l|c|c|c|c|c|} \hline
\textbf{Problem} & \textbf{All error (Recall)} & \textbf{Syntactic (Recall)} & \textbf{Strategic (Recall)} & \textbf{Conceptual (Recall)} & \textbf{Precision}\\ \hline
\texttt{caughtSpeeding} & 96 & 100 & 95 & 94 & 90\\ \hline
\texttt{redTicket} & 90 & 100 & 85 & 96 & 88\\ \hline
\texttt{countCode} & 97 & 100 & 85 & 96 & 82\\ \hline
\texttt{sum13} & 83 & 100 & 93 & 76 & 90\\ \hline
\texttt{canBalance} & 92 & 100 & 98 & 85 &92\\
\hline
\end{tabular}
\label{tab:four_problem}
\end{table*}

To further evaluate the robustness of our framework, we expanded the evaluation of our framework to four additional problems: \texttt{redTicket}, \texttt{countCode}, \texttt{sum13}, and \texttt{canBalance}, which cover topics of loops, strings, and arrays. These problems were selected to capture a range of logical errors within the constraints of the dataset and of varied complexity. Due to the time and resource limitations associated with expert evaluation, we focused on $200$ incorrect student submissions per problem. As shown in Table \ref{tab:four_problem}, the framework achieved high recall across the problems, ranging from $83$\% to $97$\%. The precision of the model for the four new problems was calculated in the range of $82\%$ to $92\%$. These recall and precision values indicate that most errors are being accurately identified. However, the model also highlights some subtrees of incorrect submissions that do not contain errors. Preliminary investigation suggests that these highlighted segments sometimes relate to code optimization or efficiency improvements. Additionally, some subtrees pertain to essential programming constructs that are necessary for correctly solving the given programming task. These observations point to future research exploring the connection between the highlighted code segments and students’ knowledge state, offering deeper insights into the progression of students' coding abilities.

\subsubsection*{Baseline Comparison}
Several tools have been designed to detect errors in student code, but most either do not provide explainable outputs, such as those based on LLMs, or rely on test cases or predefined rubrics to determine programming errors. We focus our comparisons on explainable approaches that do not require any test cases or rubrics. We developed a naive baseline model that predicts the average number of errors per problem for comparison. Using Root Mean Squared Error (RMSE) as the evaluation metric, the naive baseline achieved $1.65$, while our framework outperformed it with an RMSE of $0.45$. To compare the performance of our approach further, we considered another explainable model, \texttt{code2vec}~\cite{alon2019code2vec,shi2021toward}, which splits ASTs into different code paths using an attention network. A \texttt{code2vec} model was trained, and the code paths were extracted for the incorrect solutions to the \texttt{caughtSpeeding} problem. Due to the labor-intensive nature of expert labeling, we utilized GPT-4o-mini, inspired by recent literature using LLMs as evaluators~\cite{mcclure2024deductive,chiang2023can}. To ensure a consistent baseline comparison, GPT-4o-mini was used to evaluate the outputs of our model as well. We employed few-shot prompting, providing the LLM with examples of human-labeled data to facilitate in-context learning~\cite{brown2020language}. Based on the evaluation, \texttt{code2vec} achieved a recall of $43$\% and a precision of $37$\%. Our framework outperformed \texttt{code2vec} with a recall of $97$\% and a precision of $85$\%.

\subsection{Tracing Student Knowledge}

\begin{table*}
\centering
\caption{Performance (AUC\%) comparison for knowledge tracing. Significant improvements of SANN-DKT over DKT are marked with *($p<0.05$), **($p<0.01$), and ***($p<0.001$).}
%\vspace{2mm}
\begin{tabular}{|c|c|c|c|c|c|c|} \hline
\textbf{Model} & \textbf{A1} & \textbf{A2} & \textbf{A3} & \textbf{A4} & \textbf{A5} & \textbf{All assignments} \\ \hline
DKT & 71.24 & 73.09 & 76.84 & 69.16 & 75.14 & 65.87 \\ \hline
Code-DKT & 74.31 & 76.56 & 80.4 & 72.75 & 79.14 & - \\ \hline
SANN-DKT & 80*** & 79.08** & 79.8* & 81.2*** & 82.75** & 72.45** \\ \hline
\end{tabular}
% \begin{minipage}{\linewidth}
% \centering
%   \footnotesize 
% \end{minipage}
\label{tab:dkt}
\end{table*}

We used code vectors extracted from the modified SANN model to improve the original DKT model~\cite{piech2015deep} for knowledge tracing. We hypothesize that the extracted code vectors' implicit embedding of logical error information would contribute to tracing student knowledge and be effective in predicting future submission success. We implemented the original DKT model with an LSTM network and incorporated code vectors with the DKT inputs. We used student submissions for the 50 problems for the experiment. The train-validation-test split was 80:10:100. The split was done at the student level to avoid overlap between the different splits. We used Keras-tuner\footnote{https://keras.io/keras\_tuner/} to tune the model's hyperparameters and set the number of neurons of the LSTM model to $200$, a dropout of $0.2$, and a learning rate of $0.005$. 

We calculated the AUC score to measure the performance of our modified DKT model with code vectors in predicting students' future success in upcoming attempts to solve a problem. We compared the performance of the original DKT model against our approach by training on all of the problems. Our approach showed a higher AUC score of $72.45$\% than the original DKT model ($65.87$\%). Table \ref{tab:dkt} also shows assignment-wise AUC scores for the models, including Code-DKT~\cite{yang2022code}, which was trained for single assignments. The experimental results suggest that our approach shows significant potential in improving deep knowledge tracing, with about $10$\% higher AUC score than the original DKT. For perspective, our improvement over the original DKT ($10$\%) can be compared to Code-DKT's $3$-$4$\% improvement on CodeWorkout~\cite{yang2022code} or SAINT+'s $2.76$\% improvement on EdNet~\cite{choi2020ednet}.

\subsection{Predicting Student Performance}
We employed code embeddings from our model to predict students' final exam grades based on their programming submissions across the 50 problems in the course curriculum. Predicting a student's final exam grade necessitates comprehensive information on their programming submissions. We fed sequential code vectors to train an LSTM-based student performance prediction model to forecast final exam grades. This approach enables us to evaluate the effectiveness of the generated code vectors in encapsulating sequential insights regarding students' programming proficiency and learning trajectory. We included several baseline models to gauge our method's performance. These included a no-skill model (predicting the average of the final exam grades), Linear Regression, Lasso Regression, SVM, and Gaussian Process Regression (GPR). For the non-sequential models, we flattened the code features for all 50 problems for an individual student to accommodate the inputs. Additionally, we incorporated a baseline model (LSTM-code2vec) from prior literature~\cite{yoder2022exploring}. The hyper-parameters for the models were set as follows: for Lasso Regression, $\alpha$ was chosen from $\{1\mathrm{e}{-10}, 1\mathrm{e}{-8}, 1\mathrm{e}{-6}, 1\mathrm{e}{-4}, 1\mathrm{e}{-2}, 0.1, 1, 2, 5\}$; for SVM, the regularization parameter $C$ was selected as $1$ from the set $\{0.001, 0.1, 1, 10\}$ and the kernel was set to $``rbf"$ from $\{$$``linear"$, $``poly"$, $``rbf"$$\}$; for GPR, the kernel was set to $``rbf"$, with a length scale of $10$ selected from $\{0.5, 1, 10\}$ and noise level of $5$ from $\{0.5, 1, 5\}$. Additionally, we tuned our model based on the cross-validation results. We set the number of neurons in the LSTM model to $512$ and the number of neurons in the fully connected layer to $128$ based on the model's performance on the validation data.

\begin{table}
\centering
\caption{Students' final exam grade prediction.}
%\vspace{2mm}
\begin{tabular}{|c|c|c|} \hline
\textbf{Model} & \textbf{RMSE (std)} & \textbf{$\mathrm{R^2}$ (std)} \\ \hline
no-skill & 0.25 (0.02) & -0.01 (0.02)\\ \hline
Linear & 0.38 (0.3) & -0.27 (0.30) \\ \hline
Lasso & 0.19 (0.03) & 0.19 (0.20) \\ \hline
SVM & 0.18 (0.03) & 0.35 (0.10) \\ \hline
GPR & 0.21 (0.02) & 0.07 (0.08) \\ \hline
LSTM-code2vec & 0.23 (0.02) & 0.09 (0.16) \\ \hline
Proposed model & 0.17 (0.03) & 0.50 (0.10) \\ \hline
\end{tabular}
\label{tab:final_grade_prediction}
\end{table}

To evaluate model performance, we measured RMSE to assess the deviation between predicted and actual final exam grades. RMSE is particularly suitable as it penalizes larger errors more heavily, providing insights into model accuracy \cite{yoder2022exploring}. Additionally, we reported the coefficient of determination ($\mathrm{R^2}$), which indicates how well the independent variables explain the variation in final exam grades. Together, these metrics offer a comprehensive view of model performance. The results of student performance prediction are presented in Table \ref{tab:final_grade_prediction}. Our LSTM model outperformed other models with the lowest RMSE of $0.17$ and the highest $\mathrm{R^2}$ of $0.50$, demonstrating its superior predictive capability for student final exam grades. These findings suggest the potential of our modified SANN-based code representations in encoding not just the syntactic and semantic elements of programming but also the effectiveness in capturing the latent states of students' learning, which can offer insights into students' learning progression during the semester.

% Inefficient code patterns  

\section{Discussion}
Logical errors provide insight into students' challenges when learning and applying programming skills, including revealing potential underlying misconceptions. This insight can guide automated adaptive support to identify and address students' struggles with learning, broadening access to effective instructional support for introductory programming students. Novice programmers often struggle to identify logical errors, in part due to limited debugging experience and potential misconceptions in their programming knowledge~\cite{fitzgerald2008debugging,mccracken2001multi}. Our approach for automated identification of logical errors from students' programming submissions uses only program correctness information. This lays the foundation for building a generalized, scalable model that can reliably analyze students' programs without significant human input, such as rubrics or designed test cases. An important characteristic of our approach is its ability to explain its decision-making process, contributing to the trustworthiness of its decisions. 

Misconceptions are the most common source of logical errors \cite{ettles2018common}. Additionally, incorporating misconception information can significantly improve student modeling \cite{liu2016modeling}. We demonstrated the relationship between embedded programming code vectors implicitly encompassing students' competencies and logical errors, and students' learning by integrating students' consecutive embedded problem-solving attempts in an LSTM for short-term (next attempt) and long-term (final exam grade) success. Predicting final exam scores and predicting the next attempt success from behavioral data has been challenging in the learning analytics field ~\cite{kennedy2015predicting,yoder2022exploring,yang2022code}. We hypothesize that this is because most data used in such analysis are metadata (e.g., number of submissions, time on task, detailed rubrics), dynamic assessment of students' programming, such as test-case results ~\cite{ettles2018common}, or code context without adequate competency and misconception information. Therefore, there is not enough resemblance between the data used for predicting students' performance and tracking students' learning progression. On the other hand, vectors that are directly trained on students' programs and in accordance with their correctness stand a better chance of capturing similar information about students' understanding of programming concepts and skills as tested through programming questions on the final exam~\cite{liu2016modeling}. 

In this paper, we focused on identifying logical errors and demonstrating how incorporating logical error information enhances student modeling, specifically by improving predictive accuracy in knowledge tracing. This approach introduces new possibilities for personalized learning and intervention strategies. By leveraging information directly from students' programming submissions, these predictive models can provide educators with valuable insights into students' academic progress and performance. This information can be used to identify students at risk of academic failure and enable educators to provide targeted support and interventions to help these students succeed~\cite{yang2022code}.

While this work advances the automated identification of logical errors in students' programming submissions, it is important to acknowledge several limitations of this study. First, the dataset used in this study consisted of only Java programs, which were comparatively short in length (on average, $10-20$ statements). However, this is typical of CS1 programming skill development and practice tasks. Second, we discarded uncompilable programs due to the need for AST generation, consistent with the literature~\cite{shi2021more,shi2021toward}, resulting in the loss of a portion of our data. Third, we did not consider logical errors stemming from misinterpretation and carelessness due to the inherent difficulty in identifying these latent constructs from behavioral data. Tracing students' learning progress across multiple solutions can shed light on the underlying reasons behind their behavior. For example, if a model traces students' problem-solving behavior across multiple problems and recognizes a high probability of a student's proficiency with regard to a programming construct, a later mistake related to this construct can represent carelessness or misinterpretation. Finally, during our evaluation, we identified a small subset (less than 1\%) of incomplete submissions, lacking one or more critical components of the required functionality. This case can stem from students' carelessness or misinterpretation, or even from the trait of writing a portion of the code, submitting it, and checking whether it is correct or not. These submissions still contained some correct segments and did not necessarily exhibit logical errors in the existing code. Previous research on logical error detection has often overlooked such cases. Interestingly, our framework highlighted substructures indicative of missing constructs, suggesting the potential to extend our study. Specifically, we could integrate additional contextual information regarding core functionalities of the programming tasks rather than relying solely on dynamic test cases. This extension would allow us to detect missing components more effectively and provide personalized hints and feedback, thereby addressing a broader range of student difficulties.

\section{Design Implications}
By encoding program snippets into latent vectors, the framework we have introduced can efficiently compare student submissions, identifying similar errors while ignoring superficial differences. This can enable instructors to understand the common error patterns in a large number of student submissions~\cite{shi2021toward}. This can also enable instructors to provide targeted feedback at scale by designing feedback for a representative sample of solutions with regard to their encompassing logical errors propagated for similar solutions~\cite{silva2019adaptive}.

Beyond immediate feedback, the framework can support personalized learning by recommending targeted practice problems and instructional materials based on students’ errors. By linking incorrect submissions to a repository of similar problems, students could receive recommendations on exercises to reinforce concepts where they have struggled the most. Worked examples, particularly beneficial for novice programmers, can be suggested to highlight correct approaches and common pitfalls, further scaffolding student learning~\cite{hosseini2020improving, barria2023adaptive,hoq2025automated}. Additionally, tracking students’ problem-solving patterns allows for intelligent student modeling, enabling dynamic adjustments to learning pathways and problem difficulty~\cite{huang2023supporting}. This personalized, data-driven approach can promote deeper engagement, reduce unproductive struggles, and ensure continuous learning progression, making programming education more effective, reliable, and scalable.

% More problems
% Sloppiness and misinterpretation
% Automated correctness grading of programs
% automated process for understanding misconception
% misconception tracing from the abstract doc

%. This categorization can lead to a better understanding of the possible causes of logical errors and potential latent misconceptions \cite{albrecht2020sometimes}.%

\section{Conclusion and Future Work}
Timely identification and intervention in students'   programming struggles are crucial for enhancing engagement, learning, and, ultimately, retention. However, the persistent lack of resources to provide students with individualized instructional support poses a significant challenge in providing students with the crucial help they need while learning how to program. Automated and explainable data-driven approaches can help fill this gap by providing struggling students with adaptive support and assistance through reliable frameworks that address their individual needs. Current methods either rely heavily on manual effort to annotate the data or use complex large language models to provide support that lacks reliability and explainability. For effective integration in the classroom, we need to have code analysis methodologies that are reliable and explainable and require minimal input from human experts while still providing actionable insights for instructors and students.

This paper presented a framework for automated, explainable, and scalable identification of logical errors, leveraging their implicit presence in program embeddings to predict students' performance in programming courses. We accurately identified logical errors by training a modified SANN model on student programming submissions. Furthermore, we achieved promising results in predicting students' short-term (next problem attempts) success with $10$\% improvement over the original DKT and long-term (final exam grades) success in the course with $5$\% improvement over the best-performing baseline. Our findings demonstrate the potential of our framework in identifying student errors from their programming submissions to capture intricate details on their programming learning progression. This capability offers promise for seamless integration into adaptive instructional technologies, enabling automated and personalized scaffolding, including feedback and recommendations to support student learning experiences and outcomes.

This work suggests several promising directions for future work. First, it will be instructive to investigate how to automatically map student logical errors to underlying misconceptions by analyzing program elements and understanding the deficiencies in students' incorrect submissions. Second, exploring a misconception tracing approach that leverages multiple occurrences of related logical errors across student programs can enhance our understanding of students' learning challenges and inform targeted interventions to effectively address misconceptions. Third, integrating the framework into adaptive learning environments to provide students with personalized feedback and opportunities for mastery learning can reveal how to most effectively improve student learning experiences and outcomes in computer science education.

%% identification of the section in the article metadata and the
%\begin{acks}
%This research was supported by the XXXXXXX under Grants XXXXXXX and XXXXXXXX. Any opinions, findings, and conclusions expressed in this material are those of the authors and do not necessarily reflect the views of the XXX.
%\end{acks}

%ACKNOWLEDGMENTS are optional
\section{Acknowledgments}
This research was supported by the National Science Foundation (NSF) under Grants DUE-2236195 and DUE-2331965. Any opinions, findings, and conclusions expressed in this material are those of the authors and do not necessarily reflect the views of the NSF.

\bibliographystyle{abbrv}
\bibliography{sigproc}  % sigproc.bib is the name of the Bibliography in this case

\begin{thebibliography}{10}

\bibitem{albrecht2020sometimes}
E.~Albrecht and J.~Grabowski.
\newblock Sometimes it's just sloppiness-studying students' programming errors and misconceptions.
\newblock In {\em Proceedings of the 51st ACM Technical Symposium on Computer Science Education}, pages 340--345, New York, NY, USA, 2020. Association for Computing Machinery.

\bibitem{alon2019code2vec}
U.~Alon, M.~Zilberstein, O.~Levy, and E.~Yahav.
\newblock code2vec: Learning distributed representations of code.
\newblock {\em Proceedings of the ACM on Programming Languages}, 3(POPL):1--29, 2019.

\bibitem{altadmri201537}
A.~Altadmri and N.~C. Brown.
\newblock 37 million compilations: Investigating novice programming mistakes in large-scale student data.
\newblock In {\em Proceedings of the 46th ACM Technical Symposium on Computer Science Education}, pages 522--527, New York, NY, USA, 2015. Association for Computing Machinery.

\bibitem{alzahrani2021common}
N.~Alzahrani and F.~Vahid.
\newblock Common logic errors for programming learners: A three-decade literature survey.
\newblock In {\em Proceedings of the 2021 ASEE Annual Conference}, pages 2--18. American Society for Engineering Education, 2021.

\bibitem{ardimento2020software}
P.~Ardimento, M.~L. Bernardi, and M.~Cimitile.
\newblock Software analytics to support students in object-oriented programming tasks: An empirical study.
\newblock {\em IEEE Access}, 8:132171--132187, 2020.

\bibitem{baffes1994learning}
P.~Baffes.
\newblock Learning to model students: Using theory refinement to detect misconceptions.
\newblock Technical report, Artificial Intelligence, University of Texas at Austin, 1994.

\bibitem{baker2008more}
R.~S.~d. Baker, A.~T. Corbett, and V.~Aleven.
\newblock More accurate student modeling through contextual estimation of slip and guess probabilities in bayesian knowledge tracing.
\newblock In {\em Proceedings of the 9th International Conference on Intelligent Tutoring Systems}, pages 406--415. Springer, 2008.

\bibitem{barria2023adaptive}
J.~Barria-Pineda, K.~Akhuseyinoglu, and P.~Brusilovsky.
\newblock Adaptive navigational support and explainable recommendations in a personalized programming practice system.
\newblock In {\em Proceedings of the 34th ACM Conference on Hypertext and Social Media}, pages 1--9, New York, NY, USA, 2023. Association for Computing Machinery.

\bibitem{bayman1988using}
P.~Bayman and R.~E. Mayer.
\newblock Using conceptual models to teach basic computer programming.
\newblock {\em Journal of Educational Psychology}, 80(3):291--298, 1988.

\bibitem{bian2018nar}
P.~Bian, B.~Liang, W.~Shi, J.~Huang, and Y.~Cai.
\newblock Nar-miner: Discovering negative association rules from code for bug detection.
\newblock In {\em Proceedings of the 26th ACM Joint Meeting on European Software Engineering Conference and Symposium on the Foundations of Software Engineering}, pages 411--422, New York, NY, USA, 2018. Association for Computing Machinery.

\bibitem{brown2020language}
T.~Brown, B.~Mann, N.~Ryder, M.~Subbiah, J.~D. Kaplan, P.~Dhariwal, A.~Neelakantan, P.~Shyam, G.~Sastry, A.~Askell, et~al.
\newblock Language models are few-shot learners.
\newblock In {\em Proceedings of the 34th International Conference on Neural Information Processing Systems}, volume~33 of {\em NIPS '20}, pages 1877--1901, 2020.

\bibitem{chiang2023can}
C.-H. Chiang and H.-y. Lee.
\newblock Can large language models be an alternative to human evaluations?
\newblock {\em arXiv preprint arXiv:2305.01937}, 2023.

\bibitem{choi2020ednet}
Y.~Choi, Y.~Lee, D.~Shin, J.~Cho, S.~Park, S.~Lee, J.~Baek, C.~Bae, B.~Kim, and J.~Heo.
\newblock Ednet: A large-scale hierarchical dataset in education.
\newblock In {\em Proceedings of the International Conference on Artificial Intelligence in Education}, pages 69--73. Springer, 2020.

\bibitem{corbett1994knowledge}
A.~T. Corbett and J.~R. Anderson.
\newblock Knowledge tracing: Modeling the acquisition of procedural knowledge.
\newblock {\em User Modeling and User-adapted Interaction}, 4:253--278, 1994.

\bibitem{davies2015using}
R.~Davies, R.~Nyland, J.~Chapman, and G.~Allen.
\newblock Using transaction-level data to diagnose knowledge gaps and misconceptions.
\newblock In {\em Proceedings of the 5th International Conference on Learning Analytics and Knowledge}, pages 113--117, New York, NY, USA, 2015. Association for Computing Machinery.

\bibitem{demszky2023can}
D.~Demszky, J.~Liu, H.~C. Hill, D.~Jurafsky, and C.~Piech.
\newblock Can automated feedback improve teachers’ uptake of student ideas? evidence from a randomized controlled trial in a large-scale online course.
\newblock {\em Educational Evaluation and Policy Analysis}, 46(3):483--505, 2024.

\bibitem{edwards2017codeworkout}
S.~H. Edwards and K.~P. Murali.
\newblock Codeworkout: Short programming exercises with built-in data collection.
\newblock In {\em Proceedings of the 2017 ACM Conference on Innovation and Technology in Computer Science Education}, pages 188--193, New York, NY, USA, 2017. Association for Computing Machinery.

\bibitem{elmadani2012data}
M.~Elmadani, M.~Mathews, and A.~Mitrovic.
\newblock Data-driven misconception discovery in constraint-based intelligent tutoring systems.
\newblock In {\em Proceedings of the 20th International Conference on Computers in Education}, pages 1--8. University of Canterbury, CSSE, 2012.

\bibitem{ettles2018common}
A.~Ettles, A.~Luxton-Reilly, and P.~Denny.
\newblock Common logic errors made by novice programmers.
\newblock In {\em Proceedings of the 20th Australasian Computing Education Conference}, pages 83--89, New York, NY, USA, 2018. Association for Computing Machinery.

\bibitem{fitzgerald2008debugging}
S.~Fitzgerald, G.~Lewandowski, R.~McCauley, L.~Murphy, B.~Simon, L.~Thomas, and C.~Zander.
\newblock Debugging: Finding, fixing and flailing, a multi-institutional study of novice debuggers.
\newblock {\em Computer Science Education}, 18(2):93--116, 2008.

\bibitem{fwa2024experience}
H.~L. Fwa.
\newblock Experience report: Identifying common misconceptions and errors of novice programmers with chatgpt.
\newblock In {\em Proceedings of the 46th International Conference on Software Engineering: Software Engineering Education and Training}, pages 233--241, New York, NY, USA, 2024. Association for Computing Machinery.

\bibitem{glandorf2024temporal}
D.~Glandorf, H.~R. Lee, G.~A. Orona, M.~Pumptow, R.~Yu, and C.~Fischer.
\newblock Temporal and between-group variability in college dropout prediction.
\newblock In {\em Proceedings of the 14th International Conference on Learning Analytics and Knowledge}, pages 486--497, New York, NY, USA, 2024. Association for Computing Machinery.

\bibitem{guzman2010data}
E.~Guzm{\'a}n, R.~Conejo, and J.~G{\'a}lvez.
\newblock A data-driven technique for misconception elicitation.
\newblock In {\em Proceedings of the User Modeling, Adaptation, and Personalization: 18th International Conference, UMAP}, pages 243--254. Springer, 2010.

\bibitem{hoq2023analysis}
M.~Hoq, P.~Brusilovsky, and B.~Akram.
\newblock Analysis of an explainable student performance prediction model in an introductory programming course.
\newblock In {\em Proceedings of the 16th International Conference on Educational Data Mining}, pages 79--90, Bengaluru, India, 2023. International Educational Data Mining Society.

\bibitem{hoq2024explaining}
M.~Hoq, P.~Brusilovsky, and B.~Akram.
\newblock Explaining explainability: Early performance prediction with student programming pattern profiling.
\newblock {\em Journal of Educational Data Mining}, 16(2):115--148, 2024.

\bibitem{hoq2023sann}
M.~Hoq, S.~R. Chilla, M.~Ahmadi~Ranjbar, P.~Brusilovsky, and B.~Akram.
\newblock {SANN}: Programming code representation using attention neural network with optimized subtree extraction.
\newblock In {\em Proceedings of the 32nd ACM International Conference on Information and Knowledge Management}, pages 783--792, New York, NY, USA, 2023. Association for Computing Machinery.

\bibitem{hoq2025automated}
M.~Hoq, A.~Patil, K.~Akhuseyinoglu, P.~Brusilovsky, and B.~Akram.
\newblock An automated approach to recommending relevant worked examples for programming problems.
\newblock In {\em Proceedings of the 56th ACM Technical Symposium on Computer Science Education (SIGCSE) V. 1}, pages 527--533, New York, NY, USA, 2025. Association for Computing Machinery.

\bibitem{hoq2024detecting}
M.~Hoq, Y.~Shi, J.~Leinonen, D.~Babalola, C.~Lynch, T.~Price, and B.~Akram.
\newblock Detecting chatgpt-generated code submissions in a cs1 course using machine learning models.
\newblock In {\em Proceedings of the 55th ACM Technical Symposium on Computer Science Education}, page 526–532, New York, NY, USA, 2024. Association for Computing Machinery.

\bibitem{hoq2024towards}
M.~Hoq, J.~Vandenberg, B.~Mott, J.~Lester, N.~Norouzi, and B.~Akram.
\newblock Towards attention-based automatic misconception identification in introductory programming courses.
\newblock In {\em Proceedings of the 55th ACM Technical Symposium on Computer Science Education (SIGCSE) V. 2}, pages 1680--1681, New York, NY, USA, 2024. Association for Computing Machinery.

\bibitem{hosseini2020improving}
R.~Hosseini, K.~Akhuseyinoglu, P.~Brusilovsky, L.~Malmi, K.~Pollari-Malmi, C.~Schunn, and T.~Sirki{\"a}.
\newblock Improving engagement in program construction examples for learning python programming.
\newblock {\em International Journal of Artificial Intelligence in Education}, 30(2):299--336, 2020.

\bibitem{hovey2019survey}
C.~L. Hovey, L.~Barker, and V.~Nagy.
\newblock Survey results on why cs faculty adopt new teaching practices.
\newblock In {\em Proceedings of the 50th ACM Technical Symposium on Computer Science Education}, pages 483--489, New York, NY, USA, 2019. Association for Computing Machinery.

\bibitem{hristova2003identifying}
M.~Hristova, A.~Misra, M.~Rutter, and R.~Mercuri.
\newblock Identifying and correcting java programming errors for introductory computer science students.
\newblock {\em ACM SIGCSE Bulletin}, 35(1):153--156, 2003.

\bibitem{huang2023supporting}
Y.~Huang, P.~Brusilovsky, J.~Guerra, K.~Koedinger, and C.~Schunn.
\newblock Supporting skill integration in an intelligent tutoring system for code tracing.
\newblock {\em Journal of Computer Assisted Learning}, 39(2):477--500, 2023.

\bibitem{1_jamjoom2021early}
M.~Jamjoom, E.~Alabdulkreem, M.~Hadjouni, F.~Karim, and M.~Qarh.
\newblock Early prediction for at-risk students in an introductory programming course based on student self-efficacy.
\newblock {\em Informatica}, 45(6):1--9, 2021.

\bibitem{jia2024assessing}
Q.~Jia, J.~Cui, R.~Xi, C.~Liu, P.~Rashid, R.~Li, and E.~Gehringer.
\newblock On assessing the faithfulness of llm-generated feedback on student assignments.
\newblock In {\em Proceedings of the 17th International Conference on Educational Data Mining}, pages 491--499. International Educational Data Mining Society, 2024.

\bibitem{jia2023attentive}
Z.~Jia, W.~Su, J.~Liu, and W.~Yue.
\newblock Attentive q-matrix learning for knowledge tracing.
\newblock {\em arXiv preprint arXiv:2304.08168}, 2023.

\bibitem{kaczmarczyk2010identifying}
L.~C. Kaczmarczyk, E.~R. Petrick, J.~P. East, and G.~L. Herman.
\newblock Identifying student misconceptions of programming.
\newblock In {\em Proceedings of the 41st ACM Technical Symposium on Computer Science Education}, pages 107--111, New York, NY, USA, 2010. Association for Computing Machinery.

\bibitem{karlovvcec2012knowledge}
M.~Karlov{\v{c}}ec, M.~C{\'o}rdova-S{\'a}nchez, and Z.~A. Pardos.
\newblock Knowledge component suggestion for untagged content in an intelligent tutoring system.
\newblock In {\em Proceedings of the International Conference on Intelligent Tutoring Systems}, pages 195--200, 2012.

\bibitem{kennedy2015predicting}
G.~Kennedy, C.~Coffrin, P.~De~Barba, and L.~Corrin.
\newblock Predicting success: how learners' prior knowledge, skills and activities predict mooc performance.
\newblock In {\em Proceedings of the 5th International Conference on Learning Analytics and Knowledge}, pages 136--140, New York, NY, USA, 2015. Association for Computing Machinery.

\bibitem{ko2003development}
A.~J. Ko and B.~A. Myers.
\newblock Development and evaluation of a model of programming errors.
\newblock In {\em Proceedings of the IEEE Symposium on Human Centric Computing Languages and Environments}, pages 7--14. IEEE, IEEE, 2003.

\bibitem{landis1977measurement}
J.~R. Landis and G.~G. Koch.
\newblock The measurement of observer agreement for categorical data.
\newblock {\em Biometrics}, 33(1):159--174, 1977.

\bibitem{lee2024improving}
Y.~Lee, S.~Jeong, and J.~Kim.
\newblock Improving llm classification of logical errors by integrating error relationship into prompts.
\newblock In {\em Proceedings of the International Conference on Intelligent Tutoring Systems}, pages 91--103. Springer, 2024.

\bibitem{liang2022help}
Y.~Liang, T.~Peng, Y.~Pu, and W.~Wu.
\newblock Help-dkt: an interpretable cognitive model of how students learn programming based on deep knowledge tracing.
\newblock {\em Scientific Reports}, 12(1):4012--4023, 2022.

\bibitem{liu2016blending}
M.-h. Liu.
\newblock Blending a class video blog to optimize student learning outcomes in higher education.
\newblock {\em The Internet and Higher Education}, 30:44--53, 2016.

\bibitem{liu2016modeling}
R.~Liu, R.~Patel, and K.~R. Koedinger.
\newblock Modeling common misconceptions in learning process data.
\newblock In {\em Proceedings of the 6th International Conference on Learning Analytics and Knowledge}, pages 369--377, New York, NY, USA, 2016. Association for Computing Machinery.

\bibitem{anon_3}
J.~Marsden, S.~Yoder, and B.~Akram.
\newblock {Predicting Student Performance with Control-flow Graph Embeddings}.
\newblock In B.~Akram, T.~W. Price, Y.~Shi, P.~Brusilovsky, and S.~I. Han~Hsiao, editors, {\em 6th Educational Data Mining in Computer Science Education (CSEDM) Workshop}, pages 32--40, Durham, UK, 2022.
\newblock \url{https://doi.org/10.5281/zenodo.6983402}.

\bibitem{martin2022intelligent}
A.~C. Martin, K.~M. Ying, F.~J. Rodr{\'\i}guez, C.~S. Kahn, and K.~E. Boyer.
\newblock Intelligent support for all? a literature review of the (in) equitable design \& evaluation of adaptive pedagogical systems for cs education.
\newblock In {\em Proceedings of the 53rd ACM Technical Symposium on Computer Science Education-Volume 1}, pages 996--1002, New York, NY, USA, 2022. Association for Computing Machinery.

\bibitem{marwan2022adaptive}
S.~Marwan, B.~Akram, T.~Barnes, and T.~W. Price.
\newblock Adaptive immediate feedback for block-based programming: Design and evaluation.
\newblock {\em IEEE Transactions on Learning Technologies}, 15(3):406--420, 2022.

\bibitem{mcclure2024deductive}
J.~McClure, D.~Smyslova, A.~J. Hall, and S.~Jiang.
\newblock Deductive coding's role in ai vs. human performance.
\newblock In B.~Paassen and C.~D. Epp, editors, {\em Proceedings of the 17th International Conference on Educational Data Mining}, pages 809--813. International Educational Data Mining Society, 2024.

\bibitem{mccracken2001multi}
M.~McCracken, V.~Almstrum, D.~Diaz, M.~Guzdial, D.~Hagan, Y.~B.-D. Kolikant, C.~Laxer, L.~Thomas, I.~Utting, and T.~Wilusz.
\newblock A multi-national, multi-institutional study of assessment of programming skills of first-year cs students.
\newblock In {\em Proceedings of the Working Group Reports from ITiCSE on Innovation and Technology in Computer Science Education}, pages 125--180. Association for Computing Machinery, New York, NY, USA, 2001.

\bibitem{nathan2003expert}
M.~J. Nathan and A.~Petrosino.
\newblock Expert blind spot among preservice teachers.
\newblock {\em American Educational Research Journal}, 40(4):905--928, 2003.

\bibitem{paul2013hunting}
W.~Paul and J.~Vahrenhold.
\newblock Hunting high and low: Instruments to detect misconceptions related to algorithms and data structures.
\newblock In {\em Proceeding of the 44th ACM Technical Symposium on Computer Science Education}, pages 29--34, New York, NY, USA, 2013. Association for Computing Machinery.

\bibitem{pereyra2017regularizing}
G.~Pereyra, G.~Tucker, J.~Chorowski, {\L}.~Kaiser, and G.~Hinton.
\newblock Regularizing neural networks by penalizing confident output distributions.
\newblock {\em arXiv preprint arXiv:1701.06548}, 2017.

\bibitem{piech2015deep}
C.~Piech, J.~Bassen, J.~Huang, S.~Ganguli, M.~Sahami, L.~J. Guibas, and J.~Sohl-Dickstein.
\newblock Deep knowledge tracing.
\newblock In {\em Proceedings of the 29th International Conference on Neural Information Processing Systems}, page 505–513, Cambridge, MA, USA, 2015. MIT Press.

\bibitem{qian2017students}
Y.~Qian and J.~Lehman.
\newblock Students’ misconceptions and other difficulties in introductory programming: A literature review.
\newblock {\em ACM Transactions on Computing Education (TOCE)}, 18(1):1--24, 2017.

\bibitem{raigoza2017study}
J.~Raigoza.
\newblock A study of students' progress through introductory computer science programming courses.
\newblock In {\em Proceedings of the 2017 IEEE Frontiers in Education Conference (FIE)}, pages 1--7. IEEE, 2017.

\bibitem{rivers2013automatic}
K.~Rivers and K.~R. Koedinger.
\newblock Automatic generation of programming feedback: A data-driven approach.
\newblock In {\em Proceedings of the First Workshop on AI-supported Education for Computer Science (AIEDCS)}, volume~50, pages 50--59, 2013.

\bibitem{3_shahiri2015review}
A.~M. Shahiri, W.~Husain, et~al.
\newblock A review on predicting student's performance using data mining techniques.
\newblock {\em Procedia Computer Science}, 72:414--422, 2015.

\bibitem{shi2024evaluating}
Y.~Shi, T.~Barnes, M.~Chi, and T.~Price.
\newblock Evaluating multi-knowledge component interpretability of deep knowledge tracing models in programming.
\newblock In B.~Paassen and C.~D. Epp, editors, {\em Proceedings of the 17th International Conference on Educational Data Mining}, pages 288--295. International Educational Data Mining Society, 2024.

\bibitem{yang2022code}
Y.~Shi, M.~Chi, T.~Barnes, and T.~Price.
\newblock Code-dkt: A code-based knowledge tracing model for programming tasks.
\newblock In A.~Mitrovic and N.~Bosch, editors, {\em Proceedings of the 15th International Conference on Educational Data Mining}, pages 50--61. International Educational Data Mining Society, 2022.

\bibitem{shi2021more}
Y.~Shi, Y.~Mao, T.~Barnes, M.~Chi, and T.~W. Price.
\newblock More with less: Exploring how to use deep learning effectively through semi-supervised learning for automatic bug detection in student code.
\newblock In I.-H. Hsiao, S.~Sahebi, F.~Bouchet, and J.~Vie, editors, {\em Proceedings of the 14th International Conference on Educational Data Mining}, pages 446--453. International Educational Data Mining Society, 2021.

\bibitem{shi2021toward}
Y.~Shi, K.~Shah, W.~Wang, S.~Marwan, P.~Penmetsa, and T.~Price.
\newblock Toward semi-automatic misconception discovery using code embeddings.
\newblock In {\em Proceedings of the 11th International Conference on Learning Analytics and Knowledge}, pages 606--612, New York, NY, USA, 2021. Association for Computing Machinery.

\bibitem{silva2019adaptive}
P.~Silva, E.~Costa, and J.~R. de~Ara{\'u}jo.
\newblock An adaptive approach to provide feedback for students in programming problem solving.
\newblock In A.~Coy, Y.~Hayashi, and M.~Chang, editors, {\em Proceedings of the 15th International Conference on Intelligent Tutoring Systems, ITS}, pages 14--23. Springer, 2019.

\bibitem{sorva2013notional}
J.~Sorva.
\newblock Notional machines and introductory programming education.
\newblock {\em ACM Transactions Computing Education}, 13(2):1--31, 2013.

\bibitem{swidan2018programming}
A.~Swidan, F.~Hermans, and M.~Smit.
\newblock Programming misconceptions for school students.
\newblock In {\em Proceedings of the 2018 ACM Conference on International Computing Education Research}, pages 151--159, New York, NY, USA, 2018. Association for Computing Machinery.

\bibitem{wang2024deep}
D.~Wang, L.~Zhang, Y.~Zhao, Y.~Zhang, S.~Yan, and M.~Hou.
\newblock Deep knowledge tracking integrating programming exercise difficulty and forgetting factors.
\newblock In D.-S. Huang, Z.~Si, and W.~Chen, editors, {\em Proceedings of the International Conference on Intelligent Computing}, pages 192--203. Springer, 2024.

\bibitem{xu2024hallucination}
Z.~Xu, S.~Jain, and M.~Kankanhalli.
\newblock Hallucination is inevitable: An innate limitation of large language models.
\newblock {\em arXiv preprint arXiv:2401.11817}, 2024.

\bibitem{yoder2022exploring}
S.~Yoder, M.~Hoq, P.~Brusilovsky, and B.~Akram.
\newblock Exploring sequential code embeddings for predicting student success in an introductory programming course.
\newblock In B.~Akram, T.~W. Price, Y.~Shi, P.~Brusilovsky, and S.~I. Han~Hsiao, editors, {\em Proceedings of the 6th Educational Data Mining in Computer Science Education (CSEDM) Workshop}, pages 2--9, Durham, UK, 2022.

\bibitem{yudelson2013individualized}
M.~V. Yudelson, K.~R. Koedinger, and G.~J. Gordon.
\newblock Individualized bayesian knowledge tracing models.
\newblock In H.~C. Lane, K.~Yacef, J.~Mostow, and P.~Pavlik, editors, {\em Proceedings of the 16th International Conference on Artificial Intelligence in Education}, pages 171--180. Springer, 2013.

\bibitem{zhang2019novel}
J.~Zhang, X.~Wang, H.~Zhang, H.~Sun, K.~Wang, and X.~Liu.
\newblock A novel neural source code representation based on abstract syntax tree.
\newblock In {\em Proceedings of the 2019 IEEE/ACM 41st International Conference on Software Engineering (ICSE)}, pages 783--794. IEEE, 2019.

\end{thebibliography}
%\balancecolumns

\balancecolumns
% That's all folks!
\end{document}